\documentclass{article}

\usepackage[nonatbib, preprint]{neurips_2021}

\usepackage[utf8]{inputenc} 
\usepackage[T1]{fontenc}    
\usepackage{hyperref}       
\usepackage{xurl}
\usepackage{booktabs}       
\usepackage{amsfonts}       
\usepackage{nicefrac}       
\usepackage{microtype}      
\usepackage{xcolor}         
\usepackage{amsmath}
\usepackage[pdftex]{graphicx}
\usepackage{multirow}

\newcommand{\cut}[1]{}
\newcommand{\alfred}[1]{\textcolor{red}{[Alfred: #1]}}

\newcommand{\newparts}[1]{{#1}}

\title{On Robustness of Lane Detection Models to Physical-World Adversarial Attacks in \\ Autonomous Driving}

\author{%
  Takami Sato \\
  Department of Computer Science\\
  University of California, Irvine\\
  \texttt{takamis@uci.edu} \\
  \And
  Qi Alfred Chen \\
  Department of Computer Science\\
  University of California, Irvine\\
  \texttt{alfchen@uci.edu} \\
}

\begin{document}

\maketitle

\begin{abstract}
After the 2017 TuSimple Lane Detection Challenge, its evaluation based on accuracy and F1 score has become the de facto standard to measure the performance of lane detection methods. In this work, we conduct the first large-scale empirical study to evaluate the robustness of state-of-the-art lane detection methods under physical-world adversarial attacks in autonomous driving. We evaluate 4 major types of lane detection approaches with the conventional evaluation and end-to-end evaluation in autonomous driving scenarios, and then discuss the security proprieties of each lane detection model. 
We demonstrate that the conventional evaluation fails to reflect the robustness in end-to-end autonomous driving scenarios. Our results show that the most robust model on the conventional metrics is the least robust in the end-to-end evaluation. Although the competition dataset and its metrics have played a substantial role in developing performant lane detection methods along with the rapid development of deep neural networks, the conventional evaluation is becoming obsolete and the gap between the metrics and practicality is critical. We hope that our study will help the community make further progress in building a more comprehensive framework to evaluate lane detection models.
\end{abstract}


\section{Introduction} \label{sec:introduction}


Lane detection is an essential technology for realizing autonomous driving.
For lane detection, camera is the most frequently used sensor because it is a natural choice as lane lines are visual patterns~\cite{hillel2014recent}. Like most other computer vision areas, lane detection has been significantly benefited from the recent advances of deep neural networks (DNNs). 
In the 2017 TuSimple Lane Detection Challenge~\cite{tusimple}, DNN-based lane detection shows substantial performance as all top 3 teams apply for DNN-based lane detection. After this competition, its dataset and evaluation method based on accuracy and F1 Score became the de facto standard in lane detection evaluation. These metrics are inherited by the following datasets~\cite{pan2018spatial, llamas2019}.

However, the validity of this evaluation method in practical context, i.e., whether this is representative of practicality in real-world applications, has not been adequately researched. 
As its name implies, the main real-world applications of lane detection are for autonomous driving, e.g., online detection for automated lane centering, and offline detection for high-definition map creation, which is also mainly for high-level driving automation~\cite{TANG2021107623}. With such an application domain as its main target, the robustness of lane detection is highly critical as errors from it could be fatal.
Motivated by such high criticality, we conduct the first large-scale empirical study to evaluate the robustness of lane detection methods against physical-world adversarial attacks in autonomous driving.
We first taxonomize state-of-the-art DNN-based lane detection models into 4 major categories (\S\ref{sec:lane_detection}) and discuss the fundamental limitations of the conventional evaluation to measure the practicality for autonomous driving, especially for Automated Lane Centring (ALC), Level-2 driving automation which automatically steers a vehicle to keep it centered in the traffic lane~\cite{sae2018} (\S\ref{sec:problem}). We then introduce state-of-the-art physical-world adversarial attacks against ALC systems (\S\ref{sec:alc}).
In~\S\ref{sec:methodology}, we construct a methodology to fairly evaluate the robustness of lane detection models under physical-world adversarial attacks in the conventional accuracy evaluation and the end-to-end evaluation. For the end-to-end evaluation, we develop a bridge between lane detection methods and the vehicle lateral control implemented in OpenPilot~\cite{openpilot}, an open-source production ALC system.
In~\S\ref{sec:experiment}, we evaluate the robustness of 4 major types of lane detection approaches against 3 types of physical-world adversarial attacks. 
Throughout this study, we find that the conventional evaluation does not reflect the robustness in end-to-end autonomous driving scenarios. Our results show that the most robust model on the conventional metrics can actually be the least robust in the end-to-end evaluation. 
We then discuss the security property of each lane detection and the limitations of our study in~\S\ref{sec:discussion}.

While the TuSimple Challenge dataset and its evaluation metrics have played a substantial role in developing performant lane detection methods, the conventional evaluation method is becoming obsolete. We thus want to inform the community of the limitations of the conventional evaluation and facilitate research to build a more comprehensive evaluation methodology for lane detection since such a gap between evaluation metrics and practicality may prevent sound improvement of lane detection methods. 
All our evaluation codes will be publicly available when this work is published.

\textbf{Contributions.} Our contributions are the following: \textbf{(a)} We are the first to conduct a large-scale empirical study to measure the robustness of 4 major types of lane detection models in end-to-end autonomous driving scenarios. 
\textbf{(b)} We identify that the conventional accuracy and F1 score-based evaluation does not reflect the robustness in end-to-end autonomous driving scenarios.
\textbf{(c)} We design a methodology to fairly evaluate 4 major types of lane detection models under state-of-the-art physical-world adversarial attacks (\S\ref{sec:methodology}).
\textbf{(d)} We highlight and discuss a critical gap between conventional evaluation and practicality for autonomous driving. 






\section{Background}

\subsection{DNN-based Lane Detection} \label{sec:lane_detection}

We taxonomize state-of-the-art DNN-based lane detection methods into 4 approaches. Similar taxonomy is also applied in prior work~\cite{tabelini2021cvpr}.

\textbf{Segmentation approach.} Segmentation approach handles lane detection as a segmentation task, which classifies whether each pixel is on a lane line or not. Since this approach demonstrates the state-of-the-art performance in the 2017 TuSimple Lane Detection Challenge~\cite{tusimple} as all top-3 winners adopt the segmentation approach~\cite{pan2018spatial, hsu2018learning, neven2018towards}, this approach has been applied in many recent lane detection methods~\cite{zheng2020resa, hou2019learning}. This segmentation approach is also used in the industry. A reverse engineering study reveals that Tesla Model S applies this segmentation-based approach~\cite{jing2021tencent}.
The major drawback of this approach is its higher computational cost than the other approaches. Due to the nature of the segmentation approach, it needs to predict the classification results for every pixel, the majority of which is just background. Additionally, this approach requires a postprocessing step to extract the lane line curves from the pixel-wise classification result.

\textbf{Row-wise classification approach.} This approach~\cite{qin2020ultra, yoo2020end, hou2020inter} leverages the domain-specific knowledge that the lane lines should locate the longitudinal direction of driving vehicles and should not be so curved to have more than 2 intersections in each row of the input image. Based on the assumption, this approach formulates the lane detection task as multiple row-wise classification tasks, i.e., only one pixel per row should have a lane line. Although it still needs to output classification results for every pixel similar to the segmentation approach, this divide-and-conquer strategy enables to reduce the model size and computation while keeping high accuracy. For example, \cite{qin2020ultra} reports that their method can work at more than 300 frames per second with a comparable accuracy 95.87\% on TuSimple Challenge dataset~\cite{tusimple}. On the other hand, SAD~\cite{hou2019learning}, a segmentation approach, works at 75 frames per second with 96.64\% accuracy.
This approach also requires a postprocessing step to extract the lane lines similar to the segmentation approach.

\textbf{Curve-fitting approach.}
The curve-fitting approach~\cite{tabelini2021polylanenet, philion2019fastdraw} fits the lane lines into parametric curves (e.g., polynomials and splines). This approach is applied in an open-source production driver assistance system, OpenPilot~\cite{openpilot}. The main advantage of this approach is computationally lightweight as  OpenPilot can run on a smartphone-like device without GPU. 
To achieve lightweight computation, the accuracy is not high as other approaches. Additionally, a prior work mentions that this approach is biased toward straight lines because the majority of lane lines in the training data are straight~\cite{tabelini2021polylanenet}.

\textbf{Anchor-based approach.}
Anchor-based approach~\cite{tabelini2021cvpr, li2019line} is inspired by region-based object detectors such as Faster R-CNN~\cite{ren2015faster}. In this approach, each lane line is represented as a straight proposal line (anchor) and lateral offsets of the proposal line. Similar to the row-wise classification approach, this approach takes advantage of the domain-specific knowledge that the lane lines are generally straight. This design enables to achieve state-of-the-art latency and performance. 
LaneATT~\cite{tabelini2021cvpr} reports that it can achieve better F1 score (96.77\%) than the segmentation approaches (95.97\%) ~\cite{hou2019learning, pan2018spatial} on the TuSimple Challenge dataset.

\subsection{Limitations of Current Lane Detection Evaluation Metrics} \label{sec:problem}

All lane detection methods we discuss in~\S\ref{sec:lane_detection} evaluate their lane detection with the \textit{accuracy} and \textit{F1 score} metrics used in the 2017 TuSimple Challenge~\cite{tusimple}. The \textit{accuracy} is calculated by $\sum_{i \in H} \frac{{\rm tp}_{i}}{|H|}$, where $H$ is a set of sampled y-axis points in the driver's view image and ${\rm tp}_{i}$ is 1 if the difference of a predicted lane line point and the ground truth point at $y=i$ is within 20 pixels, otherwise 0. The detected lane line is associated with a ground truth line with the highest accuracy.
The \textit{F1 score} is a common metric to measure the performance of binary classification tasks. This is the harmonic mean of precision and recall: $\frac{2}{{\rm recall}^{-1} + {\rm precision}^{-1}}$. In the TuSimple Challenge, the precision and recall are calculated at the lane line level; The precision is the true positive ratio of detected lane lines and the recall is the true positive ratio of ground truth lines. The true positive is defined if the accuracy of a pair of the ground truth line and detected line is $\geq$ 0.85.
Although the \textit{accuracy} and \textit{F1 score} can measure a certain level capability of lane detection methods, these metrics do not fully represent the performance and practicality of the real-world applications, e.g., online detection for autonomous driving and offline detection for high-definition map creation~\cite{TANG2021107623}. 

To evaluate the practicality for autonomous driving, the evaluation based on accuracy and F1 score-based has 3 major limitations:
\textbf{(1)} There is no justification of the 20-pixel and 0.85-accuracy thresholds. For example, the ALC system can keep at the lane center as long as the detected lane lines are \textit{parallel} with actual lane lines even if the detection error is more than 20 pixels. Furthermore, the importance of detected lane line points should not be equal, i.e., the closer points to the vehicle should be more important than the distanced points to control a vehicle.
\textbf{(2)} The current metrics do not distinguish the types of lane lines: ego lane's left line, ego lane's right line, left lane's left line, etc. For the ALC system, the correct detection of the ego lane's left and right lines is critical to know the lane center. If it misdetects the left lane's left line as the ego lane's left line, the vehicle will largely deviate to the left. 
\textbf{(3)} The current metrics do not evaluate the model robustness. 
As autonomous driving is a security and safety-critical system, the model robustness is a primal factor. Typically, the model robustness and performance is a trade-off. Thus, the current evaluation may have a risk to overestimate a model overfitting to a particular dataset or its test data.

To assess the impact of the 3 limitations, we conduct an empirical study in~\S\ref{sec:experiment}. For limitation (1) and (2), we conduct an end-to-end evaluation by integrating  lane detection methods with an open-source ALC system, OpenPilot~\cite{openpilot}. For limitation (3), we evaluate the robustness of each lane detection model under 3 types of adversarial attacks.

\subsection{Physical-world Attacks for Automated Lane Centering System} \label{sec:alc}

After researchers found DNN models generally vulnerable to adversarial examples or adversarial attacks~\cite{Szegedy2014, goodfellow2014explaining}, the following work further explored such attacks in the physical world~\cite{kurakin2016adversarial_b, sharif2016accessorize, athalye2018synthesizing, brown2017advpatch, chen2018shapeshifter, eykholt2018physical, jack2018caraml, zhao2018seeing, pei2017deepxplore, tian2018deeptest, chernikova2019self, zhou2018deepbillboard}. 
Recent studies demonstrate that ALC systems, Level-2 driving automation, are also vulnerable to physical-world adversarial attacks.

\textbf{Dirty Road Patch Attack.}
Dirty Road Patch (DRP) attack is proposed as a domain-specific adversarial attack to DNN-based ALC systems~\cite{sato2020hold}. DRP attack pretends to be a benign but dirty road patch. The dirty surface pattern is generated by a white-box optimization-based method to work as an adversarial example to lane detection models. To mimic a road patch, the DRP attack has stealthiness constraints such as the gray-scale color restriction and perturbable area ratio. While it has high attack success rates, DRP attack requires white-box access to the target system and relatively heavy deployment effort.

\textbf{Drawing-Lane-Line Attack.}
As the nature of lane detection, drawing a line on the road can be an effective attack vector. A recent work~\cite{jing2021tencent} demonstrates that they can mislead Tesla Model S to the adjacent lane by putting several small stickers on the road without the original lane line. Phantom attack~\cite{nassi2020phantom} also demonstrates that they can mislead Tesla Model S by projecting fake lane lines from a drone in the nighttime. 
The drawing-lane-line attack is not as effective as the DRP attack based on our experience, but its vulnerability to this attack is more severe because of its ease of deployability.

\section{Methodology}\label{sec:methodology}

The primal goal of this study is to evaluate the gap between the conventional accuracy and F1 score-based evaluation and the practicality for autonomous driving. 
To address the limitations we discussed in~\S\ref{sec:problem}, we design a methodology to fairly evaluate the robustness of all 4 types of lane detection approaches under 3 major physical-world adversarial attacks in the conventional accuracy evaluation and the end-to-end evaluation.


\subsection{Attack Implementation}

We implemented 3 types of state-of-the-art physical-world adversarial attacks based on prior attacks against ALC systems discussed in~\ref{sec:alc}. Due to the page limit, detail of each attack implementation is in Appendix~\ref{appendx:attack_detail}.

\textbf{White-Box DRP Attack} We implement the DRP attack~\cite{sato2020hold}. While the original DRP attack uses the lane bending objective function, we apply a newly-designed 
attack objective introduced in~\S\ref{sec:objective} to conduct a fair comparison with other attacks and to deal with the output space different from the original DRP attack, which outputs detected lane lines in the bird's-eye view. All target lane detection methods in Table~\ref{tbl:lane_detection} output detected lane lines in the driver's view.

\textbf{Black-Box DRP Attack} 
To make the DRP attack work in a black-box setup, we apply a query-based black-box attack approach~\cite{ilyas2018black} to extend the DRP attack to a black-box attack. We replace the gradient calculation in the original white-box DRP attack with the gradient estimation technique NES~\cite{ilyas2018black}.

\textbf{Black-Box Drawing-Lane-Line Attack} 
We explore the most effective line with a metaheuristic strategy according to prior work~\cite{jing2021tencent}. We parameterize the drawing lane line as the start point, endpoint, and line width and optimize the parameters with the tree-structured Parzen estimator~\cite{bergstra2011algorithms} implemented in Hyperopt~\cite{hyperopt}. As the objective of the original attack~\cite{jing2021tencent} is only applicable to the segmentation approach, we optimize our original attack objective introduced in~\S\ref{sec:objective} to conduct a fair comparison with other attacks.

\subsection{Attack Objective}\label{sec:objective}

To fairly evaluate the attack capability of each attack, we formulate an attack objective function that can be commonly used for all 4 types of lane detection models. We named it the \textit{expected road center function}, which averages all detected lane lines weighted with their probabilities. Intuitively, the average of all lane lines is expected to represent the road center. If the expected center locates at the center of the input image, its value will be 0.5 in the normalized image width. We maximize the expected road center to attack to the right and minimize it to attack to the left. Detailed calculation of the expected road center for each method is in Appendix~\ref{appendx:attack_obj}. When attacking multiple frames, we average the objective of each frame over all attacking frames.

\subsection{End-to-End Simulation}\label{sec:end2end}
End-to-end robustness evaluation is an essential step in this study to highlight the gap between the conventional evaluation and the practicality for autonomous driving.
We simulate vehicle trajectories under attacks with the same methodology used in \cite{sato2020hold}. We combine a vehicle motion model~\cite{bicyclemodel} and perspective transformation~\cite{hartley2003perspective, tanaka2011perspective} to dynamically synthesize camera frame updates according to a driving trajectory. This approach enables us to evaluate the attacks on the real-world driving traces in a lightweight way.
To control a vehicle based on the lane detection results, we develop a bridge between the lane detection model and the vehicle lateral control implemented in OpenPilot~\cite{openpilot}, an open-source production ALC system. It calculates the desired driving path based on detected lane lines and makes a steering plan to follow the desired driving path with  Model Predictive Control (MPC)~\cite{simlink2020lane}. In our implementation, the desired driving path is the center of the left and right lane lines.
More details are in Appendix~\ref{appendix:openpilot}.

\textbf{Attack Goal.}
To judge the attack success in the end-to-end simulation, we follow the criteria proposed in the DRP attack~\cite{sato2020hold}. We use the attack goal achieving over 0.735 m lateral deviation on the highway within the average driver reaction, 2.5 sec. 0.735 m is the required distance to touch the lane line when a vehicle driving at the center of a 3.6m-wide highway lane. The lateral deviation is calculated between the generated trajectories with attack and without attack. Since the original human driving in the dataset sometimes does not drive at the center of the road, we compare the case with attack and without attack to more precisely measure the attack effect.
For each scenario, we consider two attack success criteria: \textit{Targeted goal} is the case that the vehicle deviates over 0.735 m to the attacking direction. \textit{Untargeted goal} is the case that the vehicle deviates over 0.735 m to either the left or right.

We also quantify the ability to drive in a benign scenario. We define a metric called \textit{benign failure rate}, which is whether the human driving and the simulated trajectory deviate by more than 0.735 m. Although the benign failure rate is expected to be always zero because ALC systems should be able to handle normal scenarios, some failure cases occur due to several reasons such as motion model inaccuracy and unstable human driving, e.g., not driving at the center of the road. 

\section{Experiments} \label{sec:experiment}

We evaluate the robustness of 4 major types of lane detection approaches against 3 adversarial attacks: white-box DRP, black-box DRP, black-box drawing-lane-line attacks. 
For each approach, we select a representative model for each approach as shown in Table~\ref{tbl:lane_detection} with the selection reasons. The pretrained weights of all models are obtained from the authors' or publicly available websites\footnote{We obtained the pretrained models from:
\begin{description}
  \setlength{\parskip}{0cm}
  \setlength{\itemsep}{0cm}
    \item[LaneATT] \url{https://github.com/lucastabelini/LaneATT}
    \item[SCNN] \url{https://github.com/harryhan618/SCNN_Pytorch}
    \item[UltraFast] \url{https://github.com/cfzd/Ultra-Fast-Lane-Detection}
    \item[PolyLaneNet] \url{https://github.com/lucastabelini/PolyLaneNet}
\end{description}
}.
All pretrained weights are training with the TuSimple Challenge training dataset~\cite{tusimple}. In all our experiments, we use a machine with the AMD Ryzen 9 3950X processor, 128GB memory, and NVIDIA RTX 3090 GPU.

\begin{table}[t]
\centering
\caption{Target lane detection methods and its selection reason. \textit{Acc.} is the accuracy of the TuSimple Challenge dataset~\cite{tusimple} in the reference papers.}
\setlength{\tabcolsep}{4pt}
\begin{tabular}{llll} \toprule
Approach & Selected Method & Acc. & Selection Reason \\ \hline
\multirow{2}{*}{Segmentaion} & \multirow{2}{*}{SCNN~\cite{pan2018spatial}} & \multirow{2}{*}{96.53\%} & \multirow{2}{*}{TuSimple Challenge winner's model} \\
 &  &  &  \\ \hline
\multirow{2}{*}{Row-wise classif.} & \multirow{2}{*}{UltraFast (ResNet18)~\cite{qin2020ultra}} & \multirow{2}{*}{95.87\%} & Highest accuracy among those whose \\
 &  &  & official code is available. \\ \hline
\multirow{2}{*}{Curve-fitting} & \multirow{2}{*}{PolyLaneNet (b0)~\cite{tabelini2021polylanenet}} & \multirow{2}{*}{88.62\%} & Highest accuracy among those whose \\
 &  &  & official code is available. \\ \hline
\multirow{2}{*}{Anchor-based} & \multirow{2}{*}{LaneATT (ResNet34)~\cite{tabelini2021cvpr}} & \multirow{2}{*}{95.63\%} & Highest accuracy among those whose \\
 &  &  & official code is available. \\ \toprule
\end{tabular}
\label{tbl:lane_detection}
\end{table}

\subsection{Conventional Evaluation Based on Accuracy and F1 Score}\label{sec:static_eval}

\begin{figure}[tbp]
\centering
\includegraphics[width=\linewidth]{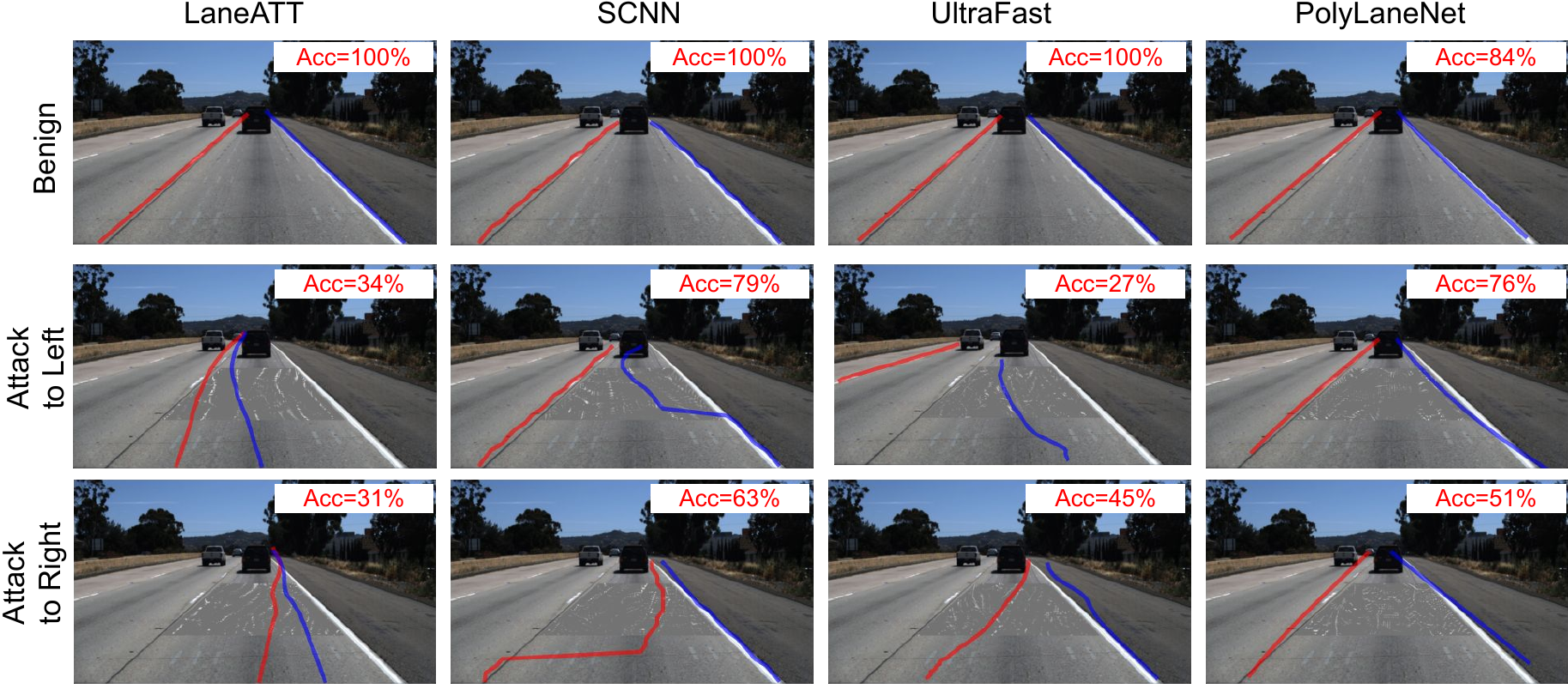}
\caption{Examples of the benign and white-box DRP attack on TuSimple Challenge data and their accuracy. The red and blue lines are the detected left and right lines respectively.
Note that the accuracy at top-right corner will not be zero if it correctly predicts that there is no lane in the sky area. }
\label{fig:poc_tusimple}
\end{figure}

\textbf{Evaluation Setup.} We first evaluate the robustness of the lane detection models with the conventional accuracy and F1 score metrics. We evaluate the lane detection models in~\S\ref{sec:lane_detection} on the TuSimple dataset\cite{tusimple}, which has 2,782 one-second-long video clips as test data. Each clip consists of 20 frames, and only the last clip is annotated and used for evaluation. We randomly select 30 clips from the test data. For each clip, we consider two attack scenarios: attack to the left, and to the right. Thus, in total, we evaluate 60 different attack scenarios. In each scenario, we place 3.6 m x 36 m patches 7 m away from the vehicle as shown in Fig.~\ref{fig:poc_tusimple}.
To deal with the limitation (2) discussed in ~\S\ref{sec:problem}, we filter out lane lines other than the ego-left and ego-right lane lines to evaluate the applicability to ALC systems more correctly. 
We thus note that the accuracy and F1 score of the benign scenarios are not consistent with prior work, which includes other lines, e.g., left lane's left line and right lane's right line. More details of each attack implementation and parameters are in Appendix~\ref{appendx:attack_detail}.

\textbf{Results.}
Table~\ref{tbl:tusimple} shows the accuracy and F1 score metrics under the 3 types of adversarial attacks: white-box DRP, black-box DRP, and black-box drawing-lane-line attacks. In the benign scenarios, the accuracy is dropped from the reported number listed in Table~\ref{tbl:lane_detection}. This indicates that the ego lane's lines are more difficult to detect correctly than other lane lines. Nevertheless, the LaneATT has only a slight decrease from 95.63\% to 94\%.
LaneATT also achieves the highest accuracy and F1 score in both the benign scenarios and all attack scenarios except for the white-box DRP attack. Contrarily, UltraFast and PolyLaneNet are the least robust models under the conventional metrics in this evaluation as they have the lowest accuracy and F1 score not only in benign scenarios but also in attack scenarios.

However, when we visually look into the detected lane lines under attack, we find quite some cases suggesting vastly different conclusions to the ones above. For example, as shown in Fig.~\ref{fig:poc_tusimple}\newparts{ and Fig.~\ref{fig:tusimple1} in Appendix~\ref{appendx:more_tusimple}}, although SCNN has the highest accuracy numbers, its detected lane lines are actually heavily curved by the attack. In contrast, PolyLaneNet's detection looks the most robust among the 4 models, as the detected lane lines are generally parallel to the actual lane lines. However, its accuracy number (76\%) is actually smaller than that of SCNN (79\%) in the attack to the left scenario. Such counter-intuitive results are because of the unreasonable 20-pixel threshold as discussed in~\S\ref{sec:problem}. Hence, the conventional accuracy and F1 score-based evaluation may not be well suited to judge the robustness of lane detection model in practical driving scenarios.

\begin{table}[t!]
\centering
\caption{Accuracy and F1 scores for attack and benign cases on the TuSimple Challenge dataset. The metrics are calculated only with ego left and right lanes. The \textbf{bold} and \underline{underlined} letters mean the highest and lowest scores, respectively, among the 4 lane detection methods. The higher score means the better robustness. 
}
\label{tbl:tusimple}. 
\renewcommand{\arraystretch}{1.1}
\setlength{\tabcolsep}{2.2pt}
\begin{tabular}{lccccccccc} \toprule
 & \multicolumn{4}{c}{Accuracy} &  & \multicolumn{4}{c}{F1 Score} \\ \cline{2-5} \cline{7-10} 
 & Benign & WB DRP & BB DRP & BB Draw &  & Benign & WB DRP & BB DRP & BB Draw \\ \cline{1-5} \cline{7-10} 
LaneATT~\cite{tabelini2021cvpr} & \textbf{94\%} & 51\% & \textbf{87\%} &\textbf{78\%} &  & \textbf{88\%} & \textbf{29\%} & \textbf{77\%} & \textbf{63\%} \\
SCNN~\cite{pan2018spatial} & 89\% & \textbf{58\%} & 86\% & 72\% &  & 75\% & 28\% & 69\% & 37\% \\
UltraFast~\cite{qin2020ultra} & 87\% & \underline{36\%} & 83\% & \underline{58\%} &  & 77\% & \underline{8\%} & 72\% & \underline{35\%} \\
PolyLaneNet~\cite{tabelini2021polylanenet} & \underline{72\%} & 53\% & \underline{65\%} & 68\% &  & \underline{50\%} & 19\% & \underline{42\%} & 43\% \\ 
\toprule
\end{tabular}
\end{table}

\subsection{End-to-End Evaluation}\label{sec:end-to-end}
To evaluate the practicality for autonomous driving, we conduct an end-to-end evaluation with the methodology introduced in~\S\ref{sec:end2end}.

\textbf{Evaluation Setup.}
We collect 20 free-flow\footnote{Vehicle has at least
5-9 seconds headway.} highway driving traces from the comma2k19 dataset~\cite{comma2k19}. For each driving trace, we consider two attack scenarios: attack to the left, and to the right. Thus, in total, we evaluate 40 different attack scenarios. For the lateral control, we use OpenPilot v0.7.0. For the longitudinal control, we used the velocity in the original trace. For the motion model, we use the parameters of Toyota RAV4 2017 (e.g., wheelbase), which is used to collects the traces of the comma2k19 dataset.
We manually adjust the input image size and field-of-view to be similar to the TuSimple dataset. More details are in Appendix~\ref{appendx:input_adapt}. We use a 5.4 m x 36 m patch size, which is the same as the one used in the DRP attack~\cite{sato2020hold}. The patch is placed at 7 m away from the vehicle at the first frame. When the patch covers lane lines, we draw lane lines on the patch to keep the original lane line information. When generating the attack, we use the first 20 frames (1 second). When evaluating the attack, we use all 50 frames (2.5 seconds), the average driver's reaction time. More details of each attack implementation and parameters are in Appendix~\ref{appendx:attack_detail}.

\begin{figure}[tbp]
\begin{center}
    \includegraphics[width=\linewidth]{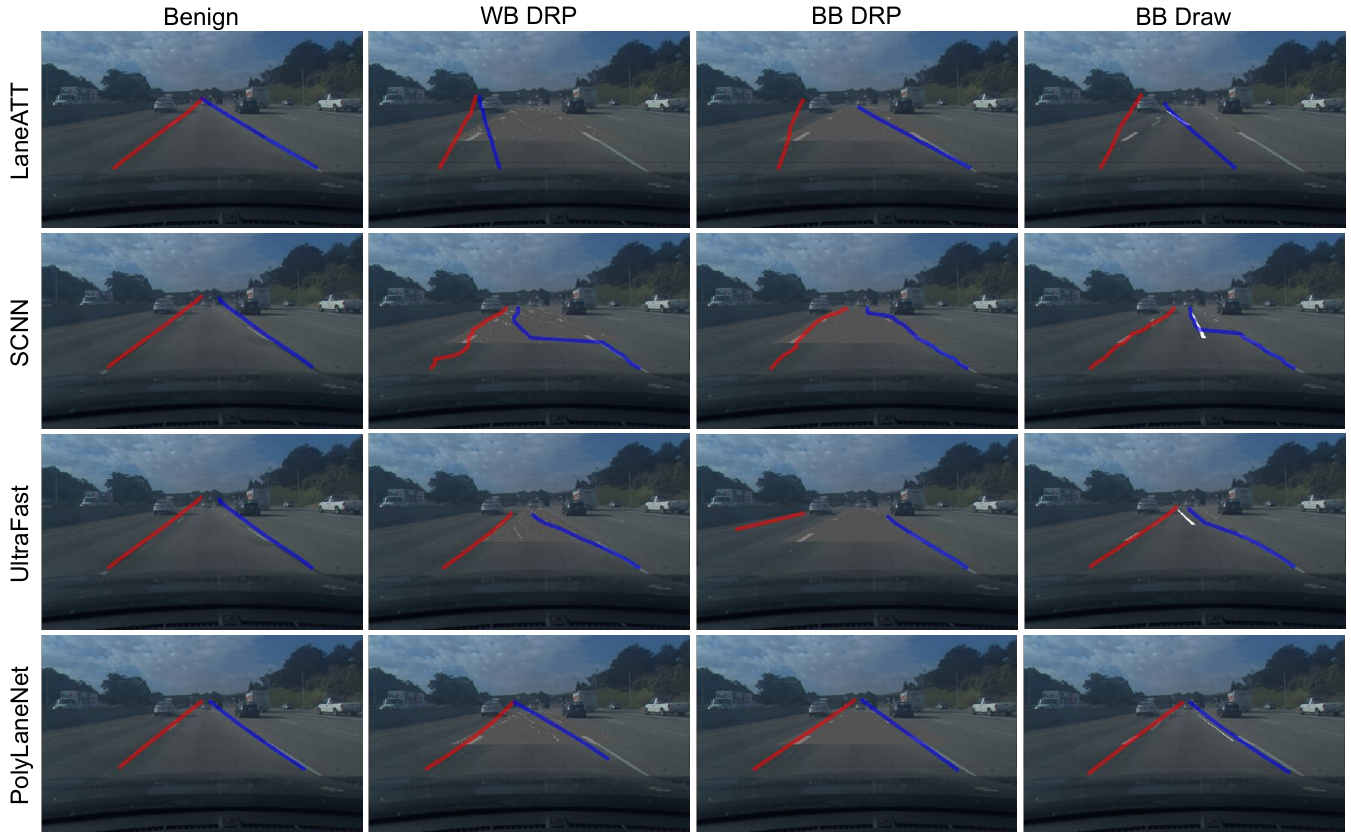}
\end{center}
\caption{
Examples of the end-to-end benign and 3 different attack scenarios on Comma2k19 data.
Each image is taken at the 4th frame (0.2 seconds after the start of the attack). The red and blue lines are the detected left and right lines respectively.}
\label{fig:poc_comma}
\end{figure}

\textbf{Results.}
Table~\ref{tbl:poc_comma} shows the results of the end-to-end evaluation. As shown, PolyLaneNet demonstrates the highest robustness as it has the lowest attack success rates in all attack scenarios. On the other hand, LaneATT, the most robust model in~\S\ref{sec:static_eval}, is the most vulnerable among the 4 lane detection methods. \textit{These results are totally the opposite of the results of conventional accuracy and F1 score evaluations in~\S\ref{sec:static_eval}.} 
In particular, LaneATT shows highly vulnerable to the black-box drawing-lane-line attack with a 90\% success rate on targeted goals and a 95\% success rate on untargeted goals.
One possibility is that the anchor-based approach is sensitive to straight lines on the road because of the design of the anchor proposal. Considering the anchor proposals are defined as straight lines, the drawing-lane-line attack may exploit the anchor representation. 
Another possibility is due to the dataset difference between the TuSimple Challenge and Comma2k19 datasets. 
As shown in Fig.~\ref{fig:poc_tusimple} and Fig.~\ref{fig:poc_comma}, the images in the TuSimple Challenge dataset typically have higher brightness and contrast than the comma2k19 dataset, thus the lane lines are more distinct. In this case, LaneATT is possibly overfitted with the TuSimple Challenge dataset. In either case, LaneATT fails to show robustness in this evaluation. Additionally, UltraFast also shows high vulnerability to the black-box drawing-lane-line attack. Fig.~\ref{fig:poc_comma} shows 4th frame (0.2 s after attack starts) of an attack scenario. All detection results except for the PolyLaneNet detection are largely changed by the attacks. The detection of SCNN is also largely changed by the black-box drawing-lane-line attack, but the attack success rate is not as high as LaneAtt and UltraFast. As can be seen from Fig.~\ref{fig:poc_tusimple} and~\ref{fig:poc_comma}, The SCNN detection tends to be parallel to the actual lane in most sections and not smoothly curved. We consider that this characteristic contributes to its robustness in end-to-end scenarios. 
In regard to the robustness, PolyLaneNet is superior to other methods and its benign failure rate is also substantially lower than the others. However, this result should be favorably influenced by PolyLaneNet's bias toward straight lines as highway roads are generally straight as discussed~\ref{sec:lane_detection}. \newparts{More details of the results in continuous frames are in Appendix~\ref{appendx:more_end2end}.}

\begin{table}[t]
\centering
\caption{Attack success rate under the end-to-end benign and 3 different attack scenarios for targeted and untargeted goals. \textit{Benign} is the benign failure rate defined in~\S\ref{sec:end-to-end}. The \textbf{bold} and \underline{underlined} letters mean the highest and lowest attack success rates, respectively, among the 4 lane detection methods. The lower attack success rate means the better robustness.
}
\label{tbl:poc_comma}
\renewcommand{\arraystretch}{1.1}
\setlength{\tabcolsep}{4pt}
\begin{tabular}{lcccccccc} \toprule
 &  & \multicolumn{3}{c}{Targeted Goal} &  & \multicolumn{3}{c}{Untargeted Goal} \\ \cline{3-5} \cline{7-9} 
 & Benign & WB DRP & BB DRP & BB Draw &  & WB DRP & BB DRP & BB Draw \\ \cline{1-5} \cline{7-9} 
LaneATT~\cite{tabelini2021cvpr} & 20\%  & \textbf{78\%} & \textbf{53\%} & \textbf{90\%} &  & \textbf{98\%} & \textbf{88\%} & \textbf{95\%} \\
SCNN~\cite{pan2018spatial} & \textbf{30\%} & \textbf{78\%} & 43\% & 58\% &  & \textbf{98\%} & 75\% & 70\% \\
UltraFast~\cite{qin2020ultra} & 25\% & 75\% & 50\% & 83\% &  & 90\% & 48\% & 93\% \\
PolyLaneNet~\cite{tabelini2021polylanenet} & \underline{5\%}  & \underline{48\%} & \underline{25\%} & \underline{30\%} &  & \underline{78\%} & \underline{43\%} & \underline{48\%} \\ \cline{1-5} \toprule
\end{tabular}
\end{table}

\begin{figure}[t]
\centering
\includegraphics[width=\linewidth]{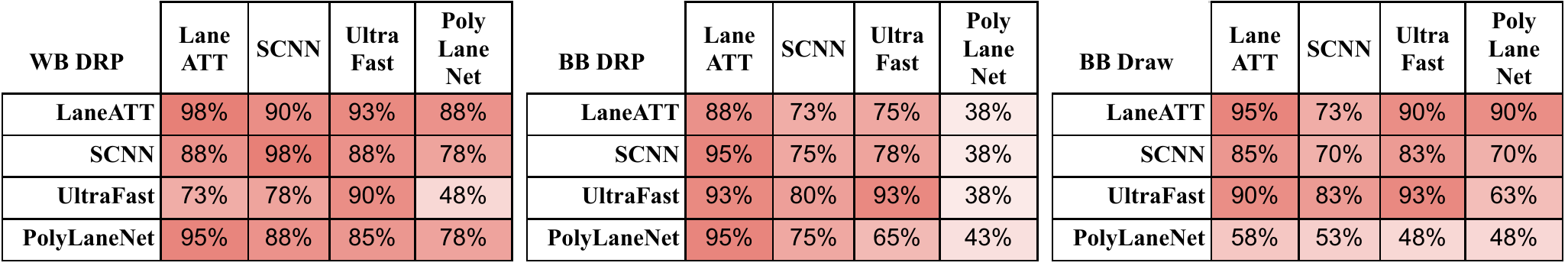}
\caption{
Transfer success rate of all pairs of models for the untargeted goal in the end-to-end scenarios. Each row indicates the source model that generates the attack and Each column indicates the target model that evaluate the patch generated with the source model.
}
\label{fig:transfer}
\end{figure}

\textbf{Transferability.} As shown in Fig.~\ref{fig:transfer}, the attack success rate is mostly less than the attack generated with target model (diagonal cells). However, the attack generated with LaneATT has high transferability to PolyLaneNet in the drawing-lane-line attack: The transfer success rate is 90\% attack success rate in the untargeted Goal.
The results indicate that PolyLaneNet also has a vulnerability to the drawing-lane-line attack, but the robustness of PolyLaneNet makes it more difficult to generate attacks.

\section{Discussion} \label{sec:discussion}


\textbf{Need to re-consider evaluation metrics for lane detection model robustness.}
The conventional accuracy and F1 score-based evaluation has critical limitations to measure the model robustness in end-to-end autonomous driving scenarios. The results of the conventional accuracy and F1 score-based evaluation are completely opposite to the end-to-end evaluation results as shown in~\S\ref{sec:experiment}. 
For safety-critical systems such as autonomous driving, it is essential to evaluate not only the performance but also the robustness. Through this study, we hope to raise the awareness that the current widely-used evaluation metrics of lane detection robustness may give a false sense of robustness in practice and the use of them could mislead the current and future efforts to improve model robustness. 



\textbf{Physical-world attack methods and model robustness.}
The white-box DRP attack has generally high attack effectiveness against all models in our evaluation. However, the black-box DRP attack shows even less effective than the black-box drawing-lane-line attack. 
One possibility is that the stealthiness constraints of the DRP attack (e.g., gray-scale color and perturbable area ratio) could be too complex to be optimized by the NES-based gradient estimation. To our knowledge, the adaptation of query-based attacks to physical-world patch attacks has not been adequately researched. 
The only effort we find is on image classifiers~\cite{feng2020query}. Thus, it is not fully studied if the query-based black-box attack is applicable for lane detection models under such complex stealthiness constraints. Further research is needed to precisely assess the security risk in black-box settings as such a naive black-box patch attack does not work.

For the drawing-lane-line attack, LaneATT~\cite{tabelini2021cvpr} shows particular vulnerability to the drawing-lane-line attack.
As discussed in~\S\ref{sec:end-to-end}, it could be due to the structure of anchor proposals in LaneATT. 
However, LaneATT is the only anchor-based method that the source code is available so far. Further research is required to confirm if the vulnerability to the drawing-lane-line attack is derived from a particular design of LaneATT or a fundamental problem of the anchor-based approach. 
Due to the ease of attack deployability, the vulnerability to the drawing-lane-line attack is severe. We thus urge the community to evaluate the robustness against naive attacks including the drawing lane line attack when designing lane detection methods.

\textbf{Defense discussions.}
As shown in~\S\ref{sec:experiment}, the white-box access to the lane detection model increases the attack success rate. Thus, the obfuscation and encryption of the model weights can have certain mitigation capabilities. However, the optimal model-level defense strategy against adversarial attacks is currently not discovered not only in the digital space but also in the physical space as mentioned in~\cite{evtimov2017physical, tramer2020, athalye2018obfuscated}. In this situation, domain-specific defenses that leverage other information available in autonomous driving are also important. One possible defense is the fusion with LiDAR and high-definition map~\cite{hdmaps}, which is a common approach used in Level-4 autonomous driving such as Google Waymo~\cite{waymoone}, although LiDAR is too costly to install commodity vehicles so far and it is needs significant efforts to correct map data. Future research needs to be conducted in both strategies, at the model level and specifically for autonomous driving.

\section{Conclusion}

In this work, we conduct the first large-scale empirical study to evaluate the robustness of 4 major types of lane detection methods under state-of-the-art 3 physical-world adversarial attacks in autonomous driving. 
We identify the fundamental limitations of the conventional evaluation to measure the practicality for autonomous driving and demonstrate that these limitations are critical to measure the model robustness in end-to-end autonomous driving scenarios: The highest robustness in the conventional evaluation shows the lowest robustness in the end-to-end evaluation. 
We also find several lane detection methods are vulnerable to the drawing-lane-line attack. The vulnerability to this attack is severe because of the ease with which attacks can be deployed. 
We thus highly recommend the community to evaluate the robustness against naive attacks such as the drawing-lane-line attack when designing lane detection methods.

In recent years, a wide variety of pretrained models have been used in many application areas such as autonomous driving~\cite{apollo}, natural language processing~\cite{bert}, and medical~\cite{chen2019med3d}. Reliable performance measurement is essential to facilitate the use of machine learning responsibly. 
The 2017 TuSimple Challenge and its accuracy and F1 score-based evaluation have played a large role to improve the performance of lane detection methods. However, in light of the fact that the recent accuracy improvement on the TuSimple dataset is only less than 1\% from the best model of the competition~\cite{laneLB}, the conventional evaluation method is ending its role. 
In autonomous driving, the underestimation of robustness hinders the sound development of lane detection models as the robustness of the model is directly related to the safety and security of autonomous driving, and can be fatal. 
Autonomous driving systems are rapidly being deployed in our society. There is an urgent demand to properly evaluate the safety and security risks of lane detection methods as early as possible.
We hope that our study will help the community make further progress in building a more comprehensive methodology to evaluate lane detection methods.

\cut{
\section{Broader Impact}
In recent years, automated driving has been rapidly implemented in our society. In automated driving, ensuring safety is the most important mission, and companies that provide automated driving services are spending a great effort for the mission~\cite{waymosafety, ubersafety}.
This study is intended to facilitate the sound development of lane detection models that reflect actual usage by demonstrating the limitations of the conventional evaluation methods of lane detection models through empirical research.
Since evaluation on a real vehicle is very costly, there is an urgent need to develop an appropriate off-line evaluation method.
We hope that this research will help the community to make further progress towards building a more comprehensive framework for evaluating lane detection models.
}

\bibliographystyle{plain}
\bibliography{main.bib}


\appendix
\section{Detailed Attack Implementation} \label{appendx:attack_detail}

In this section, we describe the detailed implementations of the 3 attacks we evaluate in this study: white-box DRP, black-box DRP, and black-box drawing-lane-line attacks. For more details on the implementation of the attacks, \newparts{we will release our source code when this work is published.}

\subsection{White-box DRP attack} \label{appendix:wb_drp}

We use the official implementation of the DRP attack~\cite{sato2020hold}. 
We obtained the source code of the DRP attack from the author and obtained permission to include the code in this submission. 
We also use parameters that are reported to have the best balance between effectiveness and secrecy: the learning rate is $10^{-2}$, regularization parameter $\lambda$ is $10^{-3}$, perturbable area ratio (PAR) is 50\%. We run 200 iterations to generate the patch in all experiments.

\subsection{Black-box DRP attack}

We use the same parameters as the white-box DRP attack shown in~\S\ref{appendix:wb_drp}: the learning rate is $10^{-2}$, regularization parameter $\lambda$ is $10^{-3}$, perturbable area ratio (PAR) is 50\%. For the parameters for the NES~\cite{ilyas2018black} gradient estimation, we generate 100 samples in each iteration, the parameter of noise magnitude $\sigma$ is $10$. The parameters of the NES refer to a popular implementation\footnote{\url{https://github.com/labsix/limited-blackbox-attacks}}.

For the number of iterations, we also used the same number as the white-box DRP attack (200 iterations) to evaluate evaluate each attack. However, the black-box attack may take more iterations to converge as the gradient information is not accurate. To evaluate the impact of the number of iterations, we test 400 iterations (200 additional iterations) on PolyLaneNet~\cite{tabelini2021polylanenet}, which shows  the highest robustness in the end-to-end evaluation (\S4.2
). The results show that the success rate increases slightly from 25\% to 27.5\% in the targeted goal, which is still highly robust compared to the other 3 attacks. The attack success rates are mostly saturated at 200 iterations. We thus think that the lower success rate of the black-box DRP attacks can be due to the complex stealthiness constraints as discussed in~\S5
.
\subsection{Black-box drawing-lane-line attack}

We reference the attack design of the prior work~\cite{jing2021tencent}. However, their implementation is dedicated to a reverse-engineered Tesla Model S setup. We thus apply several modifications to cooperate with our evaluation setup. The largest difference from the prior work is that we also explore the line angle $\theta$, which is a hyper-parameter in the prior work. 
In prior work, they try to minimize the size of drawing line and maximize the size of the detected fake lane lines instead of maximizing the vehicle deviation since they do not know how the detection results are used in the following process in Tesla.

In our implementation, we explore the line start and end points instead of deciding $\theta$. The start and end points are searched from the same area as the patch area in the DRP attack, e.g., 3.6 m x 36 m area in~\S4.1
. By this design, we can explore the line angle $\theta$ and the line length as the decision variable at the same time. The lane width is explored from 1.2 cm to 12 cm, which include a typical lane marking width, 10 cm~\cite{lane_marking_width}. 
We then decide the line start and end points and the lane width by the tree-structured Parzen estimator~\cite{bergstra2011algorithms} implemented in Hyperopt~\cite{hyperopt}. We use the same number of iterations with other attacks, 200 iterations.

\section{Details of Attack Objectives} \label{appendx:attack_obj}

To fairly evaluate the attack capability of each attack, we formulate an attack objective function, \textit{expected road center function}.
We averages all detected lane lines weighted with their probabilities.
We maximize the expected road center to attack to the right and minimize it to attack to the left. We design the expected road center function for 4 types of lane detection approaches.

\paragraph{Segmentation and row-wise classification lane detection approaches.}

For the segmentation and row-wise classification lane detection approaches, the inference results are represented as probability maps, each map being associated with a lane line. The expected center line is calculated as following:

\begin{align}
    \frac{1}{L\cdot H}\sum_{l = 1}^{L}\sum_{i = 1}^{W}\sum_{j = 1}^{H} i \cdot P^l_{ij}
\end{align}
, where $H$ and $W$ are the height and width of probability map, $L$ is the number of probability maps (channels), and $P^l_{ij}$ is the lane line existence probability of the pixel in the $(i, j)$ element of the probability map.

\paragraph{Curve-fitting approach.}

For the curve-fitting approach, the inference output is represented as the coefficients of polynomials as following:

\begin{align}
   \frac{1}{L\cdot |\mathcal{H}|}\sum_{l = 1}^{L}\sum_{j \in \mathcal{H}} [j^d, j^{d-1}, \cdots, j, 1] p_l
\end{align}
, where $L$ is the number of detected lane lines, $d$ is the degrees of polynomial ($d=3$ used in PolyLaneNet~\cite{tabelini2021polylanenet}), $\mathcal{H}$ is a set of sampled y-axis values, and $p_l\in \mathbb{R}^{d + 1}$ is the coefficient of detected lane line $l$.

\paragraph{Anchor-based approach.}

For the anchor-based approach, the inference output is represented as the probability and offsets of each anchor proposal. Thus, the expected center is obtained as following:

\begin{align}
    \sum_{l \in \mathcal{A}}
    \left [ \frac{1}{|\Delta^l|} \sum_{j \in \Delta^l}
    (a^l_j + \delta^l_j) \right ]  \cdot \pi^l
\end{align},
where $\mathcal{A}$ is a set of the anchor proposals, $\Delta^l$ is an index set of y-axis value for anchor proposal $l$, $\pi^l$ is the probability of anchor proposal $l$, and
$a^l_j$ and $\delta^l_j$ are the x-axis value and its offset of anchor proposal $l$ at y-axis index $j$ respectively.

\section{Adaptation of Comma2k19 Camera Frames into TuSimple Challenge Camera Frames} \label{appendx:input_adapt}

\begin{figure}[h]
\centering
\includegraphics[width=\linewidth]{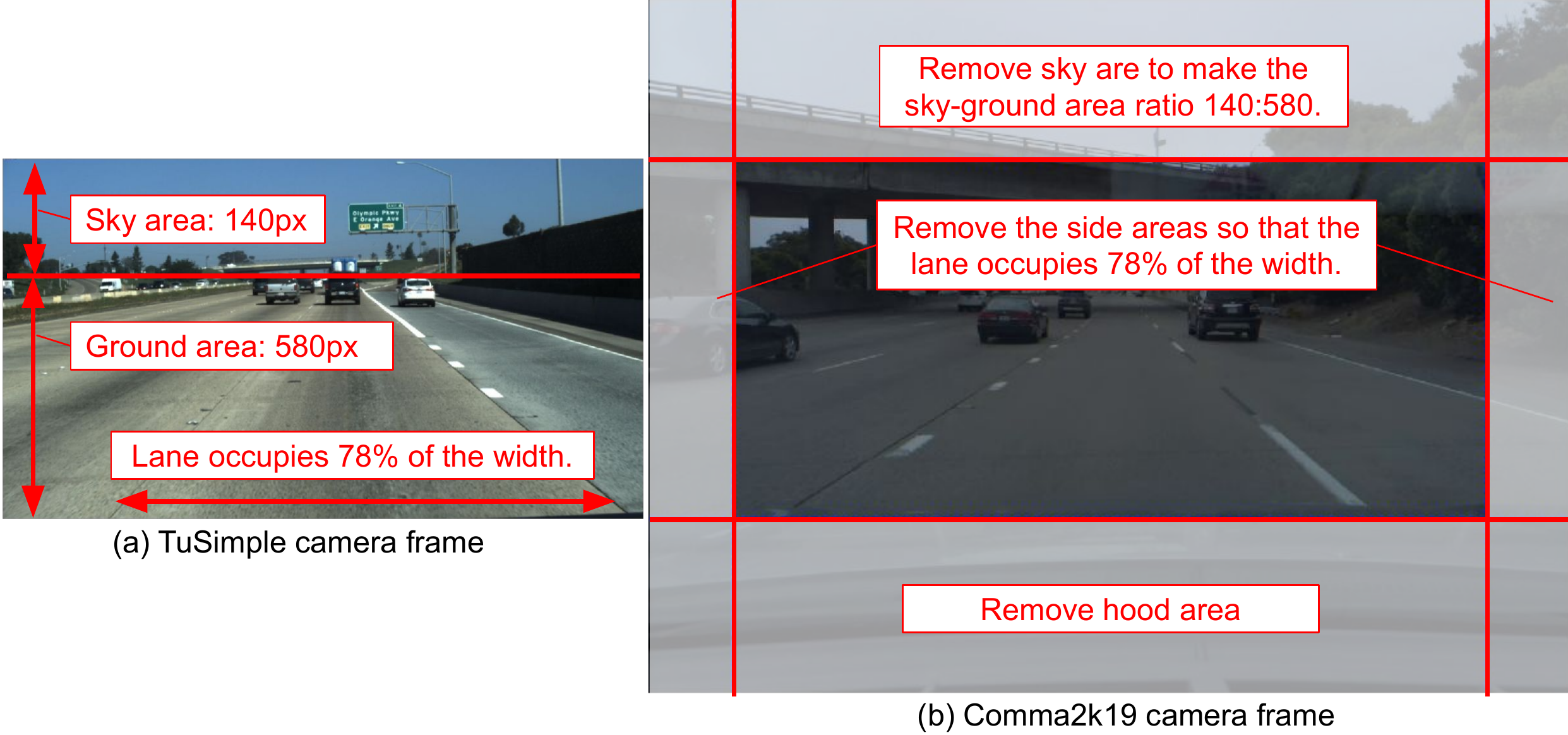}
\caption{Overview of adapting the camera frames in Comma2k19 dataset to the camera frame in the TuSimple Challenge dataset. We remove the surrounding area and use only the central part of the Comma2k19 camera frame to ensure that the comma2k19 camera frames have similar geometry as the TuSimple challenge camera frames.}
\label{fig:comma2tusimple}
\vspace{-0.2in}
\end{figure}

In the end-to-end evaluation on the comma2k19 dataset (\S4.2), we use the same pretrained models that are used in the conventional evaluation in~\S4.1, trained on the TuSimple Challenge training dataset. To deal with the differences in the datasets, we convert the camera frames in the comma2k19 dataset to have similar geometry as the camera frames in TuSimple challenge dataset.
Fig.~\ref{fig:comma2tusimple} illustrates the overview of the conversion. We remove the surrounding area and use only the central part of the Comma2k19 camera frame to have the same sky-ground area ratio and the same lane occupation ratio in the image width with the ones in the TuSimple dataset. 

\section{Details of OpenPilot ALC and its integration with lane detection models} \label{appendix:openpilot}

In this section, we explain the details of OpenPilot ALC~\cite{openpilot} and the details of its integration with the 4 lane detection models we evaluate in this study. As described in~\cite{sato2020hold}, the OpenPilot ALC system consists of 3 steps: lane detection, lateral control, and vehicle actuation.

\subsection{Lane detection} 

The image frame from the front camera is input to the lane detection model in every frame (20Hz).  Since the original OpenPilot lane detection model is a recurrent neural network model, the recurrent input from the previous frame is fed to the model with the image.  All 4 models used in this study do not have a recurrent structure, i.e., they detect lanes only in the current frame.  This is because the TuSimple Challenge has a runtime limit of less than 200 ms for each frame. Another famous dataset, CULane~\cite{pan2018spatial}, does not provide even continuous frames. In autonomous driving, the recurrent structure is a reasonable choice since past frame information is always available. Hence, the run-time calculation latency imposed in the TuSimple challenge is one of the gaps between the practicality for autonomous driving and the conventional evaluation.

\subsection{Lateral control} 

Based on the detected lane line, the lateral control decides steering angle decisions to follow the lane center (i.e., the desired driving path or waypoints) as much as possible. The original OpenPilot model outputs 3 line information: left lane line, right lane line, and driving path. The desired driving path is calculated as the average of the driving path and the center line of the left and right lane lines. The steering decision is decided by the model predictive control (MPC)~\cite{MPC}. The detected lane lines are represented in the bird's-eye-view (BEV) because the steering decision needs to be decided in a world coordinate system. 

On the contrary, all 4 models used in this study detect the lane lines in the front-camera view. We thus project the detected lane lines into the BEV space with perspective transformation~\cite{hartley2003perspective, tanaka2011perspective}. The transformation matrix for this projection is created manually based on road objects such as lane markings, and then calibrated to be able to drive in a straight lane. We create the transformation matrix for each scenario as the position of the camera and the tilt of the ground are different for each scenario. The desired driving path is calculated by the average of the left and right lane lines and fed to the MPC to decide the steering angle decisions.

In addition to the desired driving path, the MPC receives the current speed and steering angle to decide the steering angle decisions. For the steering angle, we use the human driver's steering angle in the Comma2k19 dataset in the first frame. In the following frames, the steering angle is updated by the kinematic bicycle model~\cite{bicyclemodel}, which is the most widely-used motion model for vehicles. For the vehicle speed, we use the speed of human driving in the in the comma2k19 dataset as we assume that the vehicle speed is not changed largely in the free-flow scenario, in which a vehicle has at least 5--9 seconds clear headway~\cite{boora2017identification}.

\subsection{Vehicle actuation}

The step sends steering change messages to the vehicle based on the steering angle decisions. In OpenPlot, this step operates at 100 Hz control frequency. As the lane detection and lateral control outputs the steering angle decisions in 20 Hz, the vehicle actuation sends 5 messages every steering angle decision. The  steering changes are limited to a maximum value due to the physical constraints of vehicle and for stable and for stability and safety. In this study, we limit the steering angle change to 0.25$^\circ$ following prior work, which is the steering limit for production ALC systems~\cite{sato2020hold}.

We update the vehicle states with the kinematic bicycle model based on the steering change. Note that like all motion models, the kinematic bicycle model does have approximation errors to the real vehicle dynamics~\cite{kong2015kinematic}. However, more accurate motion models require more complex parameters such as vehicle mass, tire stiffness, and air resistance~\cite{mathwork_motionmodel}. In this study, since our focus is on understanding the impact of lane detection model robustness on end-to-end driving, the most widely-used kinematic bicycle model is a sufficient choice for simulating closed-loop control behaviors. 


\section{Additional Experiment Results}

\subsection{Attack Example on TuSimple Challenge Dataset } \label{appendx:more_tusimple}

\begin{figure}[h!]
\begin{center}
    \includegraphics[width=\linewidth]{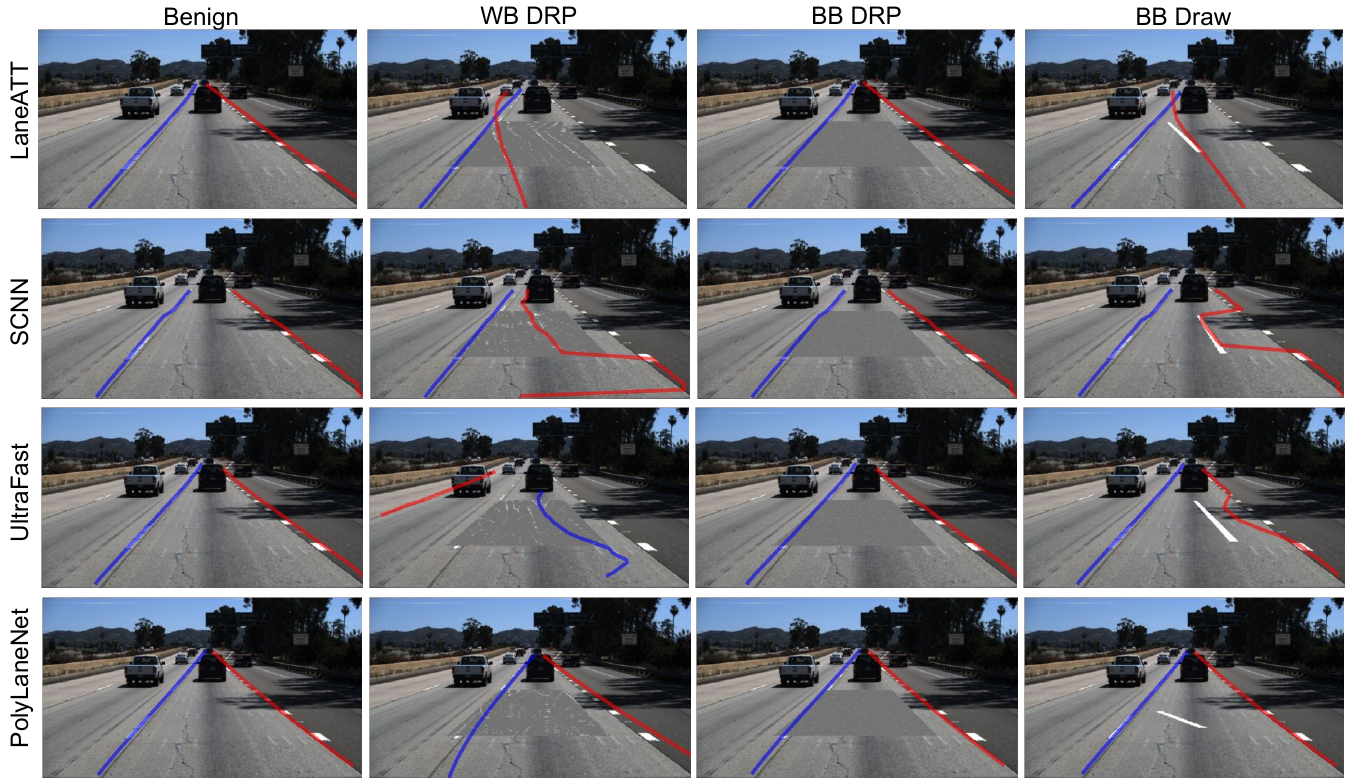}
\end{center}
\caption{
Examples of benign and 3 different attack scenarios on TuSimple Challenge data.
}
\label{fig:tusimple1}
\vspace{-0.1in}
\end{figure}

Fig.~\ref{fig:tusimple1} shows examples of benign and 3 different attack scenarios on TuSimple Challenge data. As discussed in~\S4.1, PolyLaneNet's detection looks the most robust among the 4 models, as the detected lane lines are generally parallel to the actual lane lines.

\subsection{Generated Continuous Frames in the End-to-End Simulation}\label{appendx:more_end2end}

\begin{figure}[h!]
\begin{center}
    \includegraphics[width=\linewidth]{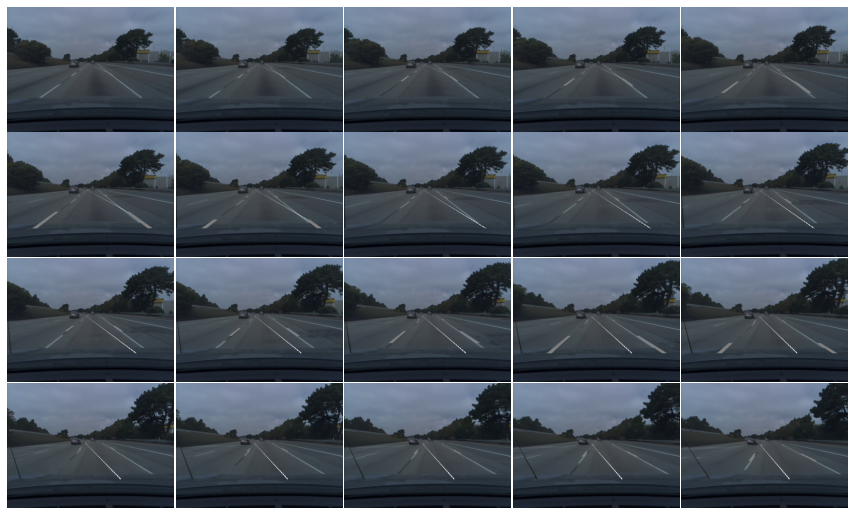}
\end{center}
\caption{
The first 20 frames (from left-top to right-bottom) of the \textbf{black-box drawing lane line attack} (to the left) on \textbf{LaneATT} in a scenario of the comma2k19 dataset. The vehicle is deviating to left due to the attack.
}
\label{fig:ex_draw_laneatt}
\vspace{0.5in}
\end{figure}

\begin{figure}[h!]
\begin{center}
    \includegraphics[width=\linewidth]{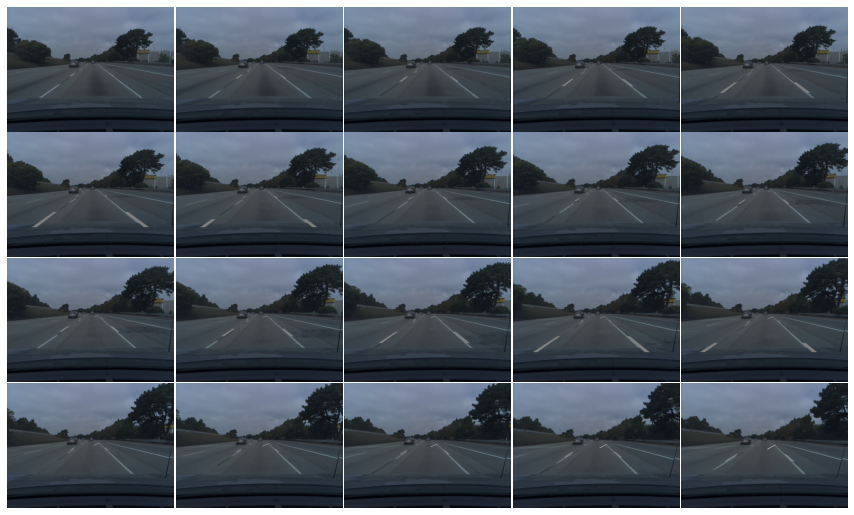}
\end{center}
\caption{
The first 20 frames (from left-top to right-bottom) of the \textbf{black-box drawing lane line attack} (to the left) on \textbf{PolyLaneNet} in a scenario of the comma2k19 dataset. The vehicle stays inside the lane even under attack.
}
\label{fig:ex_draw_poly}
\end{figure}

\begin{figure}[h!]
\begin{center}
    \includegraphics[width=\linewidth]{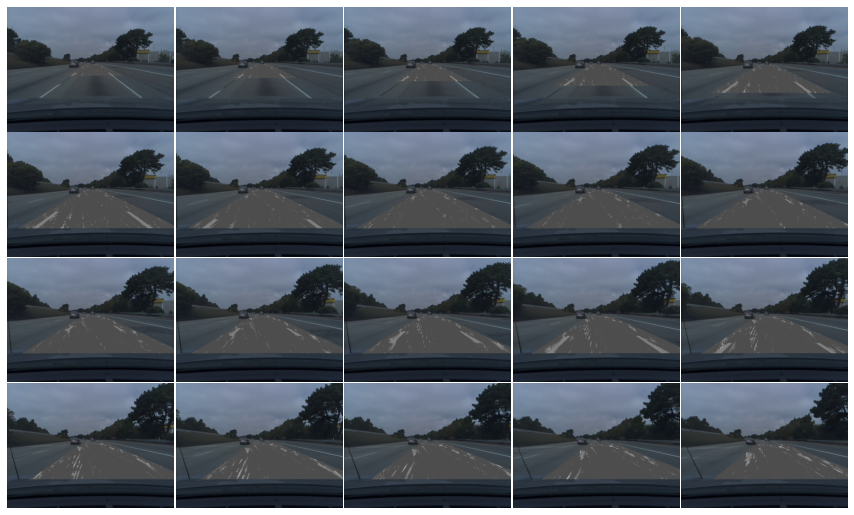}
\end{center}
\caption{
The first 20 frames (from left-top to right-bottom) of the \textbf{white-box DRP attack} (to the left) on \textbf{LaneATT} in a scenario of the comma2k19 dataset. The vehicle is deviating to left due to the attack.
}
\label{fig:ex_drp_wb_laneatt}
\vspace{0.5in}
\end{figure}

\begin{figure}[h!]
\begin{center}
    \includegraphics[width=\linewidth]{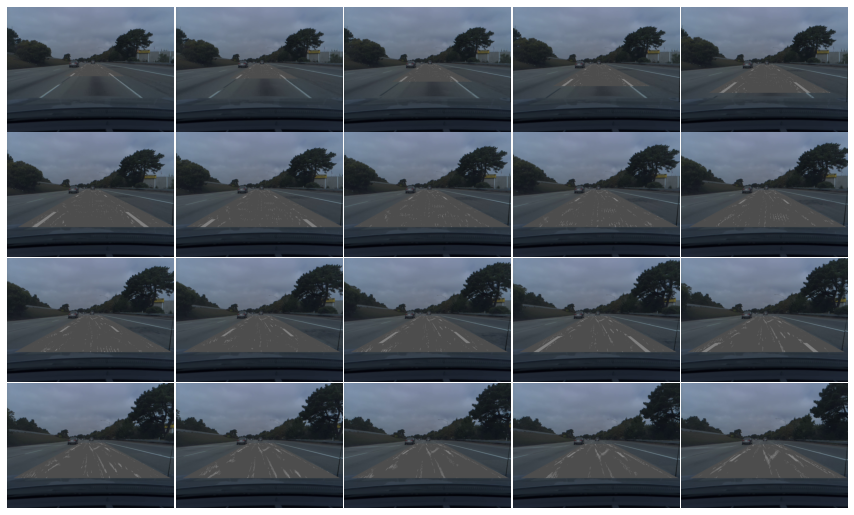}
\end{center}
\caption{
The first 20 frames (from left-top to right-bottom) of the \textbf{white-box DRP attack} (to the left) on \textbf{PolyLaneNet} in a scenario of the comma2k19 dataset. The vehicle stays inside the lane even under attack.
}
\label{fig:ex_drp_wb_poly}
\end{figure}

\begin{figure}[h!]
\begin{center}
    \includegraphics[width=\linewidth]{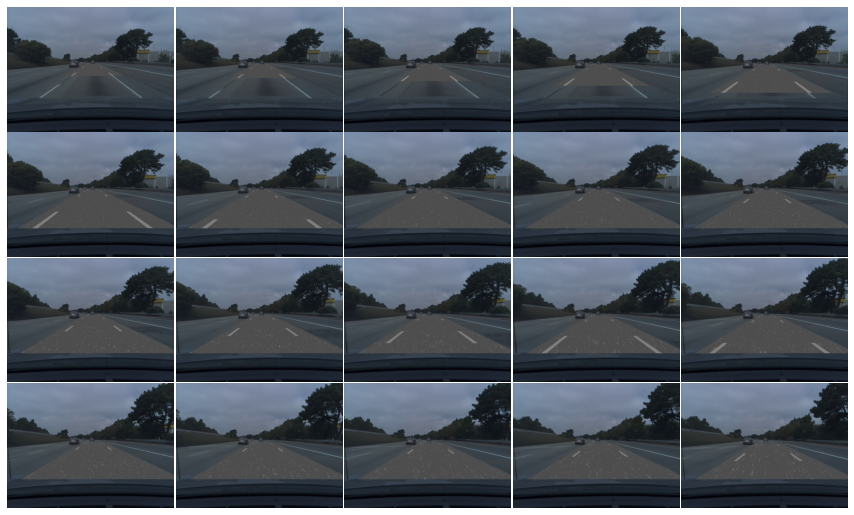}
\end{center}
\caption{
The first 20 frames (from left-top to right-bottom) of the \textbf{black-box DRP attack} (to the left) on \textbf{LaneATT} in a scenario of the comma2k19 dataset. The vehicle stays inside the lane even under attack.
}
\vspace{0.5in}
\label{fig:ex_drp_bb_laneatt}
\end{figure}

\begin{figure}[h!]
\begin{center}
    \includegraphics[width=\linewidth]{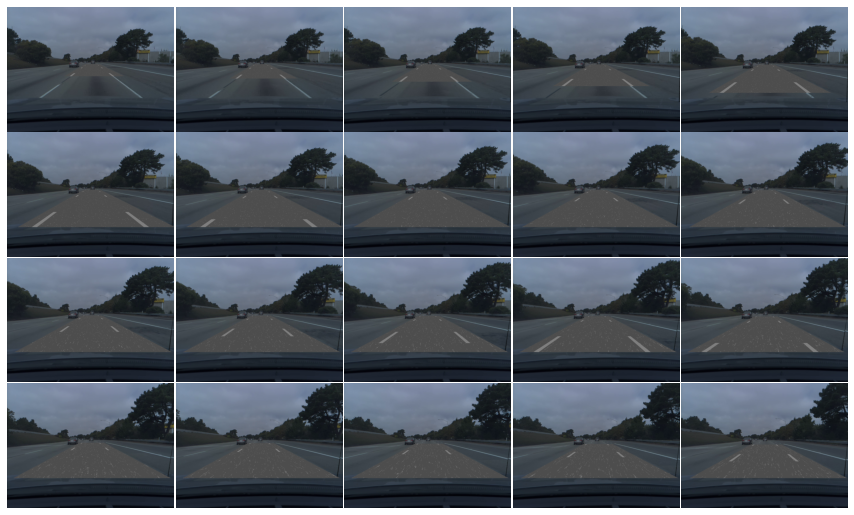}
\end{center}
\caption{
The first 20 frames (from left-top to right-bottom) of the \textbf{black-box DRP attack} (to the left) on \textbf{PolyLaneNet} in a scenario of the comma2k19 dataset. The vehicle stays inside the lane even under attack.
}
\label{fig:ex_drp_bb_poly}
\end{figure}

For the end-to-end evaluation, we synthesize font-camera frames with a vehicle motion model~\cite{bicyclemodel} and perspective transformation~\cite{hartley2003perspective, tanaka2011perspective}. Fig.~\ref{fig:ex_draw_laneatt}-\ref{fig:ex_drp_bb_poly} show the continuous frames generated with LaneATT and PolyLaneNet under the 3 attack (to the left) scenarios. 
As shown, the generated images are generally complete, with only a slight distortion in the left area as attacking to the left. We note that the distortions have almost no effect on lane detection since the side areas will be mostly removed as shown described Fig.~\ref{fig:comma2tusimple}.

Under the drawing-lane-line and white-box DRP attacks, the LaneATT drivings (in Fig.~\ref{fig:ex_draw_laneatt} and ~\ref{fig:ex_drp_wb_laneatt}) are deviated to the left and the vehicle is going out of the current lane. On the contrary, PolyLaneNet drivings (in Fig.~\ref{fig:ex_draw_poly},~\ref{fig:ex_drp_wb_poly} and~\ref{fig:ex_drp_bb_poly}) are able to stably drive within the current lane.

\end{document}